\documentclass{article}

 \usepackage[preprint]{neurips_2026}

\usepackage[utf8]{inputenc} 
\usepackage[T1]{fontenc}    
\usepackage{hyperref}       
\usepackage{url}            
\usepackage{booktabs}       
\usepackage{amsfonts}       
\usepackage{nicefrac}       
\usepackage{microtype}      
\usepackage{xcolor}         

\usepackage{amsmath,amssymb}
\usepackage{array}
\usepackage[caption=false,font=normalsize,labelfont=sf,textfont=sf]{subfig}
\usepackage{textcomp}
\usepackage{stfloats}
\usepackage{verbatim}
\usepackage{graphicx}
\usepackage{threeparttable}
\usepackage{multirow}
\usepackage{pifont}
\usepackage{algorithm}
\usepackage{algpseudocode}
\usepackage{colortbl}
\usepackage{makecell}
\usepackage{caption}
\usepackage{tabularx} 
\usepackage{subcaption}
\usepackage{wrapfig} 

\title{ViBE: Visual-to-M/EEG Brain Encoding via Spatio-Temporal VAE and Distribution-Aligned Projection}

\author{%
  Ganxi Xu$^{1}$ \quad
  Zhao-Rong Lai$^{1,*}$ \quad
  Yuting Tang$^{2,*}$ \quad
  Yonghao Song$^{5,*}$ \quad
  Shuyan Zhou$^{2,*}$ \\
  Guoxu Zhou$^{4}$ \quad
  Boyu Wang$^{3}$ \quad
  Jian Zhu$^{4}$ \quad
  Jinyi Long$^{1,\dagger}$ \\[1ex]
  $^{1}$Jinan University, Guangzhou, China \\
  $^{2}$The First Affiliated Hospital of Jinan University, Guangzhou, China \\
  $^{3}$Western University, Ontario, Canada \\
  $^{4}$Guangdong University of Technology, Guangzhou, China \\
  $^{5}$Tsinghua University, Beijing, China
}

\begin{document}

\maketitle
\begingroup
\renewcommand{\thefootnote}{}
\footnotetext{$^{*}$Equal contribution. \\ \hspace*{1.8em}$^{\dagger}$Corresponding author: \texttt{jinyil@jnu.edu.cn}}
\endgroup

\begin{abstract}
Brain encoding models not only serve to decipher how visual stimuli are transformed into neural responses, but also represent a critical step toward visual prostheses that restore vision for patients with severe vision disorders.
Brain encoding involves two fundamental steps: achieving faithful reconstruction of neural responses and establishing cross-modal alignment between visual stimuli and neural responses.
To this end, we propose ViBE, a novel brain encoding framework for generating magnetoencephalography (MEG) and electroencephalography (EEG) signals from visual stimuli.
Specifically, we first design a spatio-temporal convolutional variational autoencoder (TSC-VAE) that captures the spatio-temporal characteristics of M/EEG signals for effective neural response reconstruction.
To bridge the modality gap between visual features and neural representations, we employ Q-Former to map CLIP image embeddings to the TSC-VAE latent space, producing neural proxy embeddings.
For comprehensive cross-modal alignment, we combine mean squared error (MSE) loss for point-wise feature matching with sliced Wasserstein distance (SWD) for probability distribution alignment between the neural proxy embeddings and TSC-VAE latent embeddings.
We conduct extensive experiments on the THINGS-EEG2 and THINGS-MEG datasets, demonstrating the effectiveness of our approach in generating high-quality M/EEG signals from visual stimuli.
\end{abstract}

\section{Introduction}
\label{sec:introduction}
Brain encoding predicts neural responses from the visual stimuli that elicit them, uncovering how high-level visual semantics are represented in neural populations~\cite{naselaris2011encoding,bao2025mindsimulator,luo2023brain,luo2024brain}.
Brain encoding also corresponds to the process in visual prostheses~\cite{fernandez2018development} that utilizes a processing framework to predict neural responses for the electrode array~\cite{soltan2018head}, which has the potential to restore a rudimentary form of vision to millions of people living with incurable blindness~\cite{granley2022hybrid}.
Thus, brain encoding not only advances our understanding of human visual perception, but also lays the groundwork for applications in visual prostheses.
Researchers have been actively exploring along these two directions.
Mai \emph{et al.}~\cite{mai2025synbrain} point out that current brain encoding approaches predominantly adopt deterministic generative strategies to map visual stimuli or their semantic representations to corresponding functional magnetic resonance imaging (fMRI) recordings.
However, such deterministic models struggle with the inherent one-to-many nature of brain responses, as they yield a single fixed latent representation per input and tend to collapse diverse neural patterns into an averaged response that lacks both semantic and physiological validity.
To address this limitation, Mai \emph{et al.}~\cite{mai2025synbrain} propose SynBrain, which on one hand employs a well-designed variational model BrainVAE to generate probabilistic latent embeddings of fMRI responses that are explicitly aligned with corresponding visual semantics, and on the other hand uses mean squared error (MSE) loss to align the probabilistic latent embeddings of fMRI responses with the output features of their specifically designed Semantic-to-Neural Mapper, thereby establishing a mapping from fixed CLIP semantic embeddings to the probabilistic latent space of BrainVAE.
However, this approach introduces two problems.
(1) The probabilistic latent embeddings of fMRI recordings extracted by BrainVAE and the corresponding visual semantics extracted by the CLIP visual encoder belong to two different modalities with distinct feature scales.
Directly aligning these two types of embeddings can interfere with BrainVAE's ability to find a suitable probabilistic latent space, thereby hindering the effective reconstruction of fMRI responses.
(2) MSE loss performs point-wise feature comparison, neglecting the alignment of probability distributions between fixed CLIP semantic embeddings and probabilistic latent embeddings of fMRI responses~\cite{wang2025dataset}, which leads to distribution mismatch problems between visual semantics and corresponding fMRI recordings.
Xu \emph{et al.}~\cite{xu2025image} approach brain encoding from the perspective of visual prostheses.
Xu \emph{et al.} note that in current visual prostheses, the mainstream approach uses images as supervised signals to find suitable generated neural responses, rather than using real neural responses as supervised signals to validate the accuracy of the generated neural responses~\cite{xu2025image}.
Consequently, the limited biological resemblance of generated neural responses confines the vision restoration effect of visual prostheses to rudimentary levels~\cite{montazeri2019optogenetic}.
To address this, Xu \emph{et al.} employ the THINGS-EEG2~\cite{gifford2022large} and THINGS-MEG~\cite{hebart2023things} datasets, which contain images and their corresponding neural responses, thereby enabling the use of ground truth neural responses as supervised signals.
To this end, Xu \emph{et al.} use a diffusion transformer (DiT)~\cite{peebles2023scalable} enhanced with cross-attention mechanisms~\cite{vaswani2017attention} to achieve the conversion of images to magnetoencephalography (MEG) and electroencephalography (EEG) signals.
However, the generated M/EEG signals of this approach exhibit limited fidelity: the Pearson correlation coefficients between the generated and ground truth neural responses are only 0.425 on THINGS-EEG2 and 0.379 on THINGS-MEG~\cite{xu2025image}, indicating substantial room for improvement in the quality of brain signal generation.
To address these limitations, we propose \textbf{ViBE} (\textbf{Vi}sual-to-M/EEG \textbf{B}rain \textbf{E}ncoding), a two-stage brain encoding framework that converts visual stimuli into M/EEG signals.
Given that the high cost and low temporal resolution of fMRI recordings limit their application in brain-computer interfaces (BCIs)~\cite{li2024visual}, we focus on M/EEG signals and validate our method on the EEG dataset THINGS-EEG2 and the MEG dataset THINGS-MEG.
In Stage~I, we design a spatio-temporal convolutional variational autoencoder (TSC-VAE) that captures the hierarchical spatio-temporal structure of M/EEG signals, yielding a high-fidelity latent space as the alignment target for cross-modal mapping.
In Stage~II, we employ Q-Former~\cite{zhang2024vision} to bridge the feature scale gap between CLIP image embeddings and TSC-VAE latent embeddings, and combine MSE loss with sliced Wasserstein distance~\cite{bonneel2015sliced} to achieve both point-wise and distributional alignment.
At inference time, a test image is encoded by CLIP, projected through Q-Former into neural proxy embeddings, and decoded by TSC-VAE's decoder to produce the predicted M/EEG signals.
Our contributions are as follows:
\begin{itemize}
\item We propose a spatio-temporal convolutional variational autoencode (TSC-VAE) that fully considers the spatio-temporal characteristics of M/EEG signals, achieving effective reconstruction of M/EEG signals.
\item We employ Q-Former to map CLIP image embeddings to the same dimensional space as TSC-VAE latent embeddings, producing neural proxy embeddings that not only effectively mitigate the feature scale discrepancy between the two modalities, but also provide a foundation for cross-modal alignment.
\item We employ not only MSE loss for point-wise alignment between neural proxy embeddings and TSC-VAE latent embeddings, but also sliced Wasserstein distance for probability distribution alignment, achieving effective conversion from visual stimuli to M/EEG signals.
%
%
\end{itemize}

\section{Related Works}
\label{sec:related_works}

\subsection{Brain Encoding}
Brain encoding, which predicts neural responses from external stimuli, has been a major topic of interest in neuroscience and computational modeling~\cite{gu2022personalized,huth2016natural,mitchell2008predicting,tang2023brain}.
%
Traditional encoding approaches generally employ regression-based frameworks to map visual features to voxel-level neural responses~\cite{adeli2023predicting,beliy2024wisdom,gifford2023algonauts,luo2023brain}.
However, most methods predict only a single deterministic output, which constrains their capacity to capture neural variability.
The success of generative models has led researchers to employ them for brain encoding.
Bao \emph{et al.}~\cite{bao2025mindsimulator} leverage a diffusion-based model to learn the fMRI distribution and achieve modality alignment in a high-dimensional latent space.
Mai \emph{et al.}~\cite{mai2025synbrain} argue that such diffusion models still employ a deterministic generative process, and instead propose BrainVAE to learn a probabilistic latent space of fMRI signals conditioned on CLIP image embeddings.
However, directly aligning CLIP image embeddings with neural latent embeddings neglects the feature scale discrepancy across modalities, and all the aforementioned methods exclusively focus on visual-to-fMRI encoding, whose high cost and low temporal resolution limit their application in brain-computer interfaces~\cite{li2024visual}.
To address these issues, we propose ViBE, a novel brain encoding framework that achieves image-to-M/EEG signal conversion.
Additionally, to mitigate the scale discrepancy between CLIP image embeddings and TSC-VAE latent embeddings, we employ Q-Former to establish a mapping from CLIP image embeddings to TSC-VAE latent embeddings.

\subsection{Visual Prostheses}
Visual prostheses are neural devices designed to restore a rudimentary form of vision for patients suffering from incurable blindness caused by retinal degenerative diseases~\cite{busskamp2010genetic,cehajic2023bioengineering,prevot2020behavioural}. 
These devices typically operate through a workflow: an image sensor captures external visual scenes, a processing framework predicts stimulation patterns for the electrode array, and the implanted electrode array stimulates ganglion cells to elicit visual perception in the visual cortex~\cite{berry2017restoration,sahel2021partial,soltan2018head}.
Among these components, the processing framework that generates suitable stimulation patterns constitutes the most critical factor in determining the quality of restored vision, which has attracted substantial research efforts in recent years~\cite{turner2019stimulus}.
Existing approaches can be broadly categorized into two directions.
%
The first direction focuses on region-based detection methods that extract salient features from visual input and map them to stimulation patterns through predefined rules or learned mappings~\cite{barnich2010vibe,boyle2008region,guo2018optimization,guo2019recognition}.
%
However, the unsupervised nature of these methods means the quality of stimulation patterns cannot be verified before deployment.
%
The second direction adopts neural networks to predict stimulation patterns and employs biophysical phosphene models to simulate phosphene perception and generate decoded images~\cite{kuccukouglu2022optimization,granley2021computational,granley2022hybrid,van2024towards,van2022end}, but primarily relies on images as supervised signals without leveraging real neural responses for validation.
Xu \emph{et al.}~\cite{xu2025image} utilize the multimodal datasets THINGS-EEG2 and THINGS-MEG, which contain both images and corresponding neural responses for validation.
Nevertheless, this approach achieves only modest generation quality, with Pearson correlation coefficients of 0.425 and 0.379 on THINGS-EEG2 and THINGS-MEG, respectively~\cite{xu2025image}.
To address this issue, we combine point-wise MSE loss with sliced Wasserstein distance to jointly align neural proxy embeddings and TSC-VAE latent embeddings at both the feature and distribution levels, enabling high-fidelity conversion from visual stimuli to M/EEG signals.

\section{Methodology}
\label{sec:methodology}

\subsection{Problem Definition}
\label{sec:problem_definition}

\begin{figure*}[htbp!]
\centering
\includegraphics[width=\linewidth]{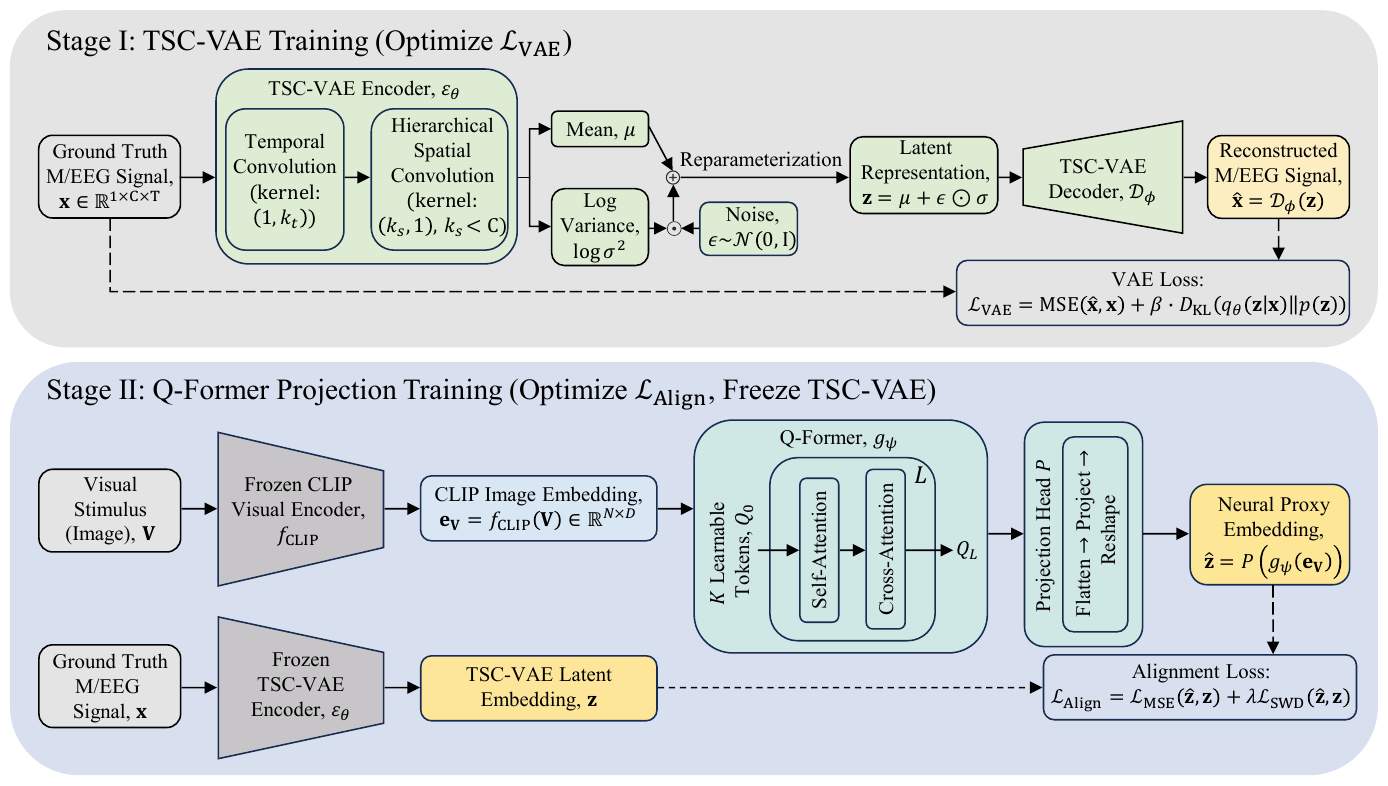}
\caption{
Overview of the training procedure. In Stage~I, the TSC-VAE employs temporal convolutions with kernel size $(1, k_t)$ and spatial convolutions with kernel size $(k_s, 1)$ where $k_s < ch$ to capture the hierarchical spatio-temporal characteristics of M/EEG signals, and is trained to reconstruct M/EEG signals using the ELBO loss. In Stage~II, Q-Former leverages learnable query tokens to attend to CLIP image embeddings via cross-attention, and the projection head $P$ produces neural proxy embeddings aligned with TSC-VAE latent embeddings through a combined MSE and SWD loss.
}
\label{fig_training}
\end{figure*}

Let $\mathbf{V}$ denote a visual stimulus (image) and $\mathbf{x} \in \mathbb{R}^{1 \times C \times T}$ denote the corresponding M/EEG neural response with $C$ channels and $T$ time points.
A pre-trained CLIP visual encoder $f_{\text{CLIP}}$ maps the image to a visual embedding $\mathbf{e}_{\mathbf{V}} = f_{\text{CLIP}}(\mathbf{V}) \in \mathbb{R}^{N \times D}$, where $N$ is the number of visual tokens and $D$ is the feature dimension.
Our goal is to learn a mapping from visual stimuli to neural responses, \emph{i.e.}, given a test image $\mathbf{V}$, generate the corresponding M/EEG signal $\hat{\mathbf{x}}$.
To achieve this, ViBE comprises two stages (Fig.~\ref{fig_training}).
In Stage~I, we train a variational autoencoder (TSC-VAE) with encoder $\mathcal{E}_{\theta}$ and decoder $\mathcal{D}_{\phi}$ that encodes M/EEG signals into a latent representation $\mathbf{z} = \mathcal{E}_{\theta}(\mathbf{x}) \in \mathbb{R}^{d \times H \times W}$ and reconstructs them, where $d$ is the number of latent channels, $H$ and $W$ are the spatial and temporal dimensions of the latent space, respectively.
In Stage~II, we train a Q-Former $g_{\psi}$ followed by a multi-layer projection head $P$ that together map CLIP image embeddings $\mathbf{e}_{\mathbf{V}}$ to neural proxy embeddings $\hat{\mathbf{z}} = P(g_{\psi}(\mathbf{e}_{\mathbf{V}})) \in \mathbb{R}^{d \times H \times W}$, which share the same shape as the TSC-VAE latent representation $\mathbf{z}$.
The predicted M/EEG signal is then obtained by decoding: $\hat{\mathbf{x}} = \mathcal{D}_{\phi}(\hat{\mathbf{z}})$.

\subsection{Stage I: Neural Response Reconstruction via TSC-VAE}
\label{sec:stage1}
Song \emph{et al.}~\cite{song2023decoding} propose TSConv, a temporal-spatial separated convolution that decomposes standard 2D convolution into a temporal convolution with kernel size $(1, k_t)$ and a spatial convolution with kernel size $(ch, 1)$, where $ch$ equals the number of channels.
However, according to the bottom-up hierarchy of the visual system, visual stimuli are processed sequentially by V1, V2, V4 on the occipital cortex, and the inferotemporal (IT) cortex on the temporal cortex along the ventral stream for object recognition~\cite{song2023decoding}.
This hierarchical processing endows M/EEG signals with distinct spatio-temporal characteristics: the spatial dimension reflects the topographic distribution of neural activity across different brain regions (channels/electrodes), while the temporal dimension captures the dynamic evolution of neural responses over time.
Setting the spatial kernel size to $(ch, 1)$ therefore collapses the entire spatial dimension in a single operation, contradicting this hierarchical nature of neural spatial information.
To address this limitation, we propose TSConvPlus, which sets the spatial kernel size to $(k_s, 1)$ with $k_s < ch$, thereby detecting the hierarchical spatial information of neural responses across different brain regions.
We build our variational autoencoder upon TSConvPlus and accordingly name it TSC-VAE (TSConvPlus-based VAE).
The TSC-VAE encoder $\mathcal{E}_{\theta}$ maps the input signal $\mathbf{x}$ to a latent distribution:
\begin{equation*}
(\boldsymbol{\mu}, \log \boldsymbol{\sigma}^2) = \mathcal{E}_{\theta}(\mathbf{x}),
\end{equation*}
where $\boldsymbol{\mu}$ and $\log \boldsymbol{\sigma}^2$ parameterize the approximate posterior distribution $q_{\theta}(\mathbf{z}|\mathbf{x})$.
The latent representation $\mathbf{z} \in \mathbb{R}^{d \times H \times W}$ is obtained via the reparameterization technique~\cite{kingma2013auto}:
\begin{equation*}
\mathbf{z} = \boldsymbol{\mu} + \boldsymbol{\epsilon} \odot \boldsymbol{\sigma}, \quad \boldsymbol{\epsilon} \sim \mathcal{N}(\mathbf{0}, \mathbf{I}).
\end{equation*}
The decoder $\mathcal{D}_{\phi}$ then reconstructs the signal as $\hat{\mathbf{x}} = \mathcal{D}_{\phi}(\mathbf{z})$.
The entire model is trained by minimizing the evidence lower bound (ELBO):
\begin{equation}
\mathcal{L}_{\text{VAE}} = \text{MSE}(\hat{\mathbf{x}}, \mathbf{x}) + \beta \cdot D_{\text{KL}}\left(q_{\theta}(\mathbf{z}|\mathbf{x}) \| p(\mathbf{z})\right),
\label{eq:vae_loss}
\end{equation}
where the first term is the reconstruction loss measured by mean squared error, $D_{\text{KL}}$ is the Kullback-Leibler divergence between the approximate posterior and the standard Gaussian prior $p(\mathbf{z}) = \mathcal{N}(\mathbf{0}, \mathbf{I})$, and $\beta$ is a weighting coefficient that balances reconstruction fidelity with latent space regularization.

\subsection{Stage II: Visual-to-Neural Mapping via Q-Former Projection}
\label{sec:stage2}
In the second stage, we address the challenge of mapping visual stimuli to neural responses.
Neural responses and images are inherently different modalities with distinct feature scales.
The CLIP image embeddings reside in a visual semantic space, while the TSC-VAE latent embeddings occupy a neural representation space.
To bridge this modality gap and reconcile the scale discrepancy, we employ Q-Former~\cite{zhang2024vision}.
Given an image $\mathbf{V}$, we first extract CLIP image embeddings using a pre-trained and frozen CLIP visual encoder~\cite{radford2021learning}:
\begin{equation*}
\mathbf{e}_{\mathbf{V}} = f_{\text{CLIP}}(\mathbf{V}) \in \mathbb{R}^{N \times D},
\end{equation*}
where $N = 257$ is the number of visual tokens (1 class token and 256 patch tokens) and $D = 768$ is the feature dimension for ViT-L/14.
The Q-Former module employs $K$ learnable query tokens $\mathbf{Q}_0 \in \mathbb{R}^{K \times d_h}$, which interact with each other via self-attention and attend to the CLIP image embeddings $\mathbf{e}_{\mathbf{V}}$ via cross-attention.
In each transformer layer $l$ ($l = 1, \ldots, L$), the query tokens undergo the following operations:
\begin{align*}
\mathbf{Q}_l^{\text{SA}} &= \text{LN}\left(\text{SelfAttn}(\mathbf{Q}_{l-1}) + \mathbf{Q}_{l-1}\right), \\
\mathbf{Q}_l^{\text{CA}} &= \text{LN}\left(\text{CrossAttn}(\mathbf{Q}_l^{\text{SA}}, \mathbf{e}_{\mathbf{V}}) + \mathbf{Q}_l^{\text{SA}}\right), \\
\mathbf{Q}_l &= \text{LN}\left(\text{FFN}(\mathbf{Q}_l^{\text{CA}}) + \mathbf{Q}_l^{\text{CA}}\right),
\end{align*}
where LN denotes layer normalization.
The self-attention layers capture inter-query dependencies, the cross-attention layers extract relevant visual information from the CLIP image embeddings, and the feed-forward network (FFN) provides additional nonlinear transformations.
After processing through all $L$ layers, the output query tokens are flattened, projected, and reshaped through a multi-layer projection head $P$ to produce the neural proxy embeddings:
\begin{equation*}
\hat{\mathbf{z}} = P(\mathbf{Q}_L) \in \mathbb{R}^{d \times H \times W},
\end{equation*}
where $\hat{\mathbf{z}}$ shares the same shape as the TSC-VAE latent representation $\mathbf{z}$.
Q-Former effectively adjusts the feature scale of CLIP image embeddings to match that of TSC-VAE latent embeddings, enabling subsequent alignment in a common dimensional space.
To achieve comprehensive alignment between the neural proxy embeddings and TSC-VAE latent embeddings, we combine MSE loss for point-wise alignment with sliced Wasserstein distance (SWD)~\cite{bonneel2015sliced} for probability distribution alignment:
\begin{equation}
\mathcal{L}_{\text{Align}} = \mathcal{L}_{\text{MSE}}(\hat{\mathbf{z}}, \mathbf{z}) + \lambda \mathcal{L}_{\text{SWD}}(\hat{\mathbf{z}}, \mathbf{z}),
\label{eq:align_loss}
\end{equation}
where $\lambda$ is a mixing parameter that controls the relative contribution of the probability distribution alignment term, and $\mathbf{z}$ is the TSC-VAE latent embedding obtained by encoding the ground truth M/EEG signal with the frozen TSC-VAE encoder.
The SWD is computed by projecting the high-dimensional embeddings onto random one-dimensional subspaces:
\begin{equation*}
\mathcal{L}_{\text{SWD}}(\hat{\mathbf{z}}, \mathbf{z}) = \frac{1}{M} \sum_{m=1}^{M} W_1\left(\boldsymbol{\theta}_m^{\top} \hat{\mathbf{z}}_{\text{flat}},\; \boldsymbol{\theta}_m^{\top} \mathbf{z}_{\text{flat}}\right),
\end{equation*}
where $\{\boldsymbol{\theta}_m\}_{m=1}^{M}$ are random projection directions sampled uniformly from the unit sphere, $\hat{\mathbf{z}}_{\text{flat}}$ and $\mathbf{z}_{\text{flat}}$ denote the flattened latent representations, and $W_1$ denotes the one-dimensional Wasserstein distance computed via sorting and mean absolute difference.
The MSE loss ensures point-wise feature correspondence, while the SWD captures distributional discrepancies that MSE alone cannot address, together providing a comprehensive alignment objective.

\subsection{Inference}
\label{sec:inference}

\begin{figure*}[htbp!]
\centering
\includegraphics[width=\linewidth]{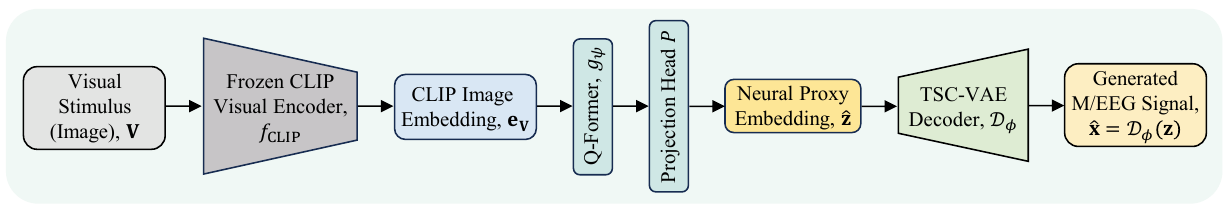}
\caption{
Illustration of the inference pipeline. Given a test image $\mathbf{V}$, the frozen CLIP visual encoder extracts image embeddings $\mathbf{e}_{\mathbf{V}}$, Q-Former with the projection head $P$ produces neural proxy embeddings $\hat{\mathbf{z}}$, and the frozen TSC-VAE decoder $\mathcal{D}_{\phi}$ generates the predicted M/EEG signal $\hat{\mathbf{x}}$.
}
\label{fig_inference}
\end{figure*}

During inference, the three components are integrated into a unified pipeline (Fig.~\ref{fig_inference}): given a test image $\mathbf{V}$, we first extract CLIP image embeddings $\mathbf{e}_{\mathbf{V}} = f_{\text{CLIP}}(\mathbf{V})$, then Q-Former produces the neural proxy embeddings $\hat{\mathbf{z}} = P(g_{\psi}(\mathbf{e}_{\mathbf{V}}))$, and finally TSC-VAE's decoder generates the predicted M/EEG signal $\hat{\mathbf{x}} = \mathcal{D}_{\phi}(\hat{\mathbf{z}})$.

\section{Experiment}
\label{sec:experiment}

\subsection{Performance Evaluation}
\label{sec:performance_evaluation}

\begin{table*}[htbp!]
\caption{Quantitative evaluation on THINGS-EEG2 and THINGS-MEG datasets.}
\label{tab_results}
\centering
\begin{minipage}[t]{0.48\textwidth}
\centering
\textbf{(a) THINGS-EEG2} \\[0.5em]
\resizebox{\textwidth}{!}{
\begin{tabular}{c|c|ccc}
\toprule
\multicolumn{2}{c|}{\multirow{2}{*}{Method}} & \multicolumn{3}{c}{Evaluation Metrics} \\
\cmidrule{3-5}
\multicolumn{2}{c|}{~} & MSE $\downarrow$ & Pearson $\uparrow$ & Cosine $\uparrow$ \\
\midrule
\multicolumn{2}{c|}{G{\"u}{\c{c}}l{\"u} \emph{et al.}~\cite{gucclu2015deep}} & 0.211 & 0.178 & 0.181 \\
\multicolumn{2}{c|}{Yamins \emph{et al.}~\cite{yamins2014performance}} & 0.196 & 0.203 & 0.209 \\
\multicolumn{2}{c|}{UNet-based Diffusion} & 0.217 & 0.188 & 0.185 \\
\multicolumn{2}{c|}{MindSimulator~\cite{bao2025mindsimulator}} & 0.160 & 0.349 & 0.345 \\
\multicolumn{2}{c|}{SynBrain~\cite{mai2025synbrain}} & 0.156 & 0.366 & 0.347 \\
\multicolumn{2}{c|}{Xu \emph{et al.}~\cite{xu2025image}} & $\boldsymbol{0.109}$ & 0.425 & 0.402 \\
\multicolumn{2}{c|}{ViBE (Ours)} & $0.954 {\pm 0.235}$ & $\boldsymbol{0.635} {\pm 0.054}$ & $\boldsymbol{0.635} {\pm 0.053}$ \\
\midrule
\multicolumn{2}{c|}{Xu \emph{et al.}~\cite{xu2025image} (Cross-subject)} & $\boldsymbol{0.174}$ & 0.157 & 0.165 \\
\multicolumn{2}{c|}{ViBE (Ours, cross-subject)} & $1.865 {\pm 0.379}$ & $\boldsymbol{0.295} {\pm 0.051}$ & $\boldsymbol{0.297} {\pm 0.049}$ \\
\midrule
\multirow{2}{*}[-4pt]{\shortstack{Leave-one-\\subject-out}} & Within-subject & $0.362 {\pm 0.025}$ & $0.480 {\pm 0.014}$ & $0.479 {\pm 0.014}$ \\
\cmidrule{2-2}
& Cross-subject & $0.396 {\pm 0.041}$ & $0.445 {\pm 0.070}$ & $0.445 {\pm 0.069}$ \\
\bottomrule
\end{tabular}
}
\label{tab_eeg_results}
\end{minipage}
\hfill
\begin{minipage}[t]{0.48\textwidth}
\centering
\textbf{(b) THINGS-MEG} \\[0.5em]
\resizebox{\textwidth}{!}{
\begin{tabular}{c|c|ccc}
\toprule
\multicolumn{2}{c|}{\multirow{2}{*}{Method}} & \multicolumn{3}{c}{Evaluation Metrics} \\
\cmidrule{3-5}
\multicolumn{2}{c|}{~} & MSE $\downarrow$ & Pearson $\uparrow$ & Cosine $\uparrow$ \\
\midrule
\multicolumn{2}{c|}{G{\"u}{\c{c}}l{\"u} \emph{et al.}~\cite{gucclu2015deep}} & 0.831 & 0.243 & 0.254 \\
\multicolumn{2}{c|}{Yamins \emph{et al.}~\cite{yamins2014performance}} & 0.796 & 0.269 & 0.274 \\
\multicolumn{2}{c|}{UNet-based Diffusion} & 0.763 & 0.122 & 0.127 \\
\multicolumn{2}{c|}{MindSimulator~\cite{bao2025mindsimulator}} & 0.755 & 0.281 & 0.274 \\
\multicolumn{2}{c|}{SynBrain~\cite{mai2025synbrain}} & 0.727 & 0.302 & 0.326 \\
\multicolumn{2}{c|}{Xu \emph{et al.}~\cite{xu2025image}} & $\boldsymbol{0.663}$ & 0.379 & 0.382 \\
\multicolumn{2}{c|}{ViBE (Ours)} & $1.451 {\pm 0.334}$ & $\boldsymbol{0.543} {\pm 0.071}$ & $\boldsymbol{0.539} {\pm 0.074}$ \\
\midrule
\multicolumn{2}{c|}{Xu \emph{et al.}~\cite{xu2025image} (Cross-subject)} & $\boldsymbol{0.768}$ & 0.082 & 0.089 \\
\multicolumn{2}{c|}{ViBE (Ours, cross-subject)} & $2.154 {\pm 0.513}$ & $\boldsymbol{0.261} {\pm 0.050}$ & $\boldsymbol{0.260} {\pm 0.050}$ \\
\midrule
\multirow{2}{*}[-4pt]{\shortstack{Leave-one-\\subject-out}} & Within-subject & $1.236 {\pm 0.368}$ & $0.449 {\pm 0.032}$ & $0.447 {\pm 0.033}$ \\
\cmidrule{2-2}
& Cross-subject & $1.415 {\pm 0.276}$ & $0.333 {\pm 0.067}$ & $0.331 {\pm 0.067}$ \\
\bottomrule
\end{tabular}
}
\label{tab_meg_results}
\end{minipage}
\end{table*}
We evaluate ViBE on the THINGS-EEG2 and THINGS-MEG datasets using three metrics: mean squared error (MSE), where lower values indicate better performance, Pearson correlation coefficient (Pearson) and cosine similarity (Cosine), where higher values indicate better performance.
We compare against six brain encoding methods: G{\"u}{\c{c}}l{\"u} \emph{et al.}~\cite{gucclu2015deep}, Yamins \emph{et al.}~\cite{yamins2014performance}, UNet-based Diffusion, MindSimulator~\cite{bao2025mindsimulator}, SynBrain~\cite{mai2025synbrain}, and Xu \emph{et al.}~\cite{xu2025image}.
The quantitative results are presented in Table~\ref{tab_results}.
%



%
We evaluate under three protocols: subject-specific training, cross-subject generalization, and leave-one-subject-out (detailed in Appendix~\ref{appendix:eval_protocols}).
Comparing the ``ViBE (Ours)'' and ``ViBE (Ours, cross-subject)'' rows in Table~\ref{tab_results}, we observe substantial performance degradation across subjects, consistent with the cross-subject variations reported in Xu \emph{et al.}~\cite{xu2025image}.
To mitigate this, we adopt a leave-one-subject-out protocol, whose cross-subject results substantially outperform the per-subject cross-subject results on both datasets.
For instance, on THINGS-EEG2, the Pearson correlation improves from 0.295 to 0.445, and on THINGS-MEG from 0.261 to 0.333, demonstrating that training on multi-subject data effectively reduces cross-subject variability.
However, the leave-one-subject-out within-subject performance falls behind the per-subject results (e.g., Pearson 0.480 vs.\ 0.635 on THINGS-EEG2).
Under the leave-one-subject-out protocol, each fold uses 148{,}860 training trials for THINGS-EEG2 and 66{,}744 for THINGS-MEG, representing a substantially larger training set.
Developing models that can fully exploit such large-scale multi-subject data remains a promising direction for future work.

\subsection{Ablation Study: TSConv vs.\ TSConvPlus}
\label{sec:ablation_conv}

\begin{figure}[htbp!]
\centering
\includegraphics[width=\textwidth]{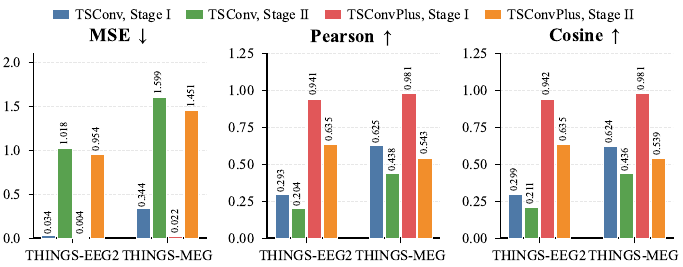}
\caption{Visual comparison of TSConv and TSConvPlus across Stage~I and Stage~II on THINGS-EEG2 and THINGS-MEG.
}
\label{fig_ablation_conv}
\end{figure}
We compare TSConvPlus with TSConv, and the results are reported in Fig.~\ref{fig_ablation_conv}.
In Stage~I, TSConvPlus achieves dramatically superior performance on both datasets. On THINGS-EEG2, TSConvPlus attains a Pearson correlation of 0.941, compared to only 0.293 for TSConv; on THINGS-MEG, the gap is similarly pronounced (0.981 vs.\ 0.625).
These results confirm that setting the spatial kernel size to $(k_s, 1)$ with $k_s < C$, rather than collapsing the entire spatial dimension with $(C, 1)$, enables the model to effectively capture the hierarchical spatial information in M/EEG signals, leading to substantially more faithful signal reconstruction.
The advantage propagates to Stage~II. On THINGS-EEG2, TSConvPlus achieves a Pearson correlation of 0.635 versus 0.204 for TSConv; on THINGS-MEG, 0.543 versus 0.438.
This demonstrates that a higher-quality latent space learned in Stage~I directly benefits the cross-modal alignment in Stage~II, as the Q-Former projects CLIP image embeddings into a more structured and informative latent space.
%
%

%
As shown in Figure~\ref{fig_ablation_conv}, a notable performance gap exists between Stage~I and Stage~II. For example, with TSConvPlus on THINGS-EEG2, the Pearson correlation drops from 0.941 (Stage~I) to 0.635 (Stage~II).
This gap reflects the inherent difficulty of cross-modal mapping. Stage~I performs within-modality autoencoding, where the encoder and decoder operate on the same neural signal space with full access to the input information. In contrast, Stage~II must map from CLIP image embeddings to neural proxy embeddings---two fundamentally different modalities.
As analyzed in Section~\ref{sec:embedding_analysis}, the CLIP image embeddings and TSC-VAE latent embeddings differ in scale by approximately two orders of magnitude. Although the Q-Former bridges roughly 40 times of this gap, a residual scale offset of approximately 2.5 times remains (Table~\ref{tab_embedding_stats}), which inherently limits Stage~II performance.
Nevertheless, the advantage of TSConvPlus over TSConv remains substantial in Stage~II (Pearson 0.635 vs.\ 0.204 on THINGS-EEG2 and 0.543 vs.\ 0.438 on THINGS-MEG), confirming that a high-fidelity latent space learned in Stage~I is critical for effective cross-modal alignment.

\subsection{Modality Gap Analysis}
\label{sec:embedding_analysis}
\begin{figure*}[htbp!]
\centering
\includegraphics[width=\textwidth]{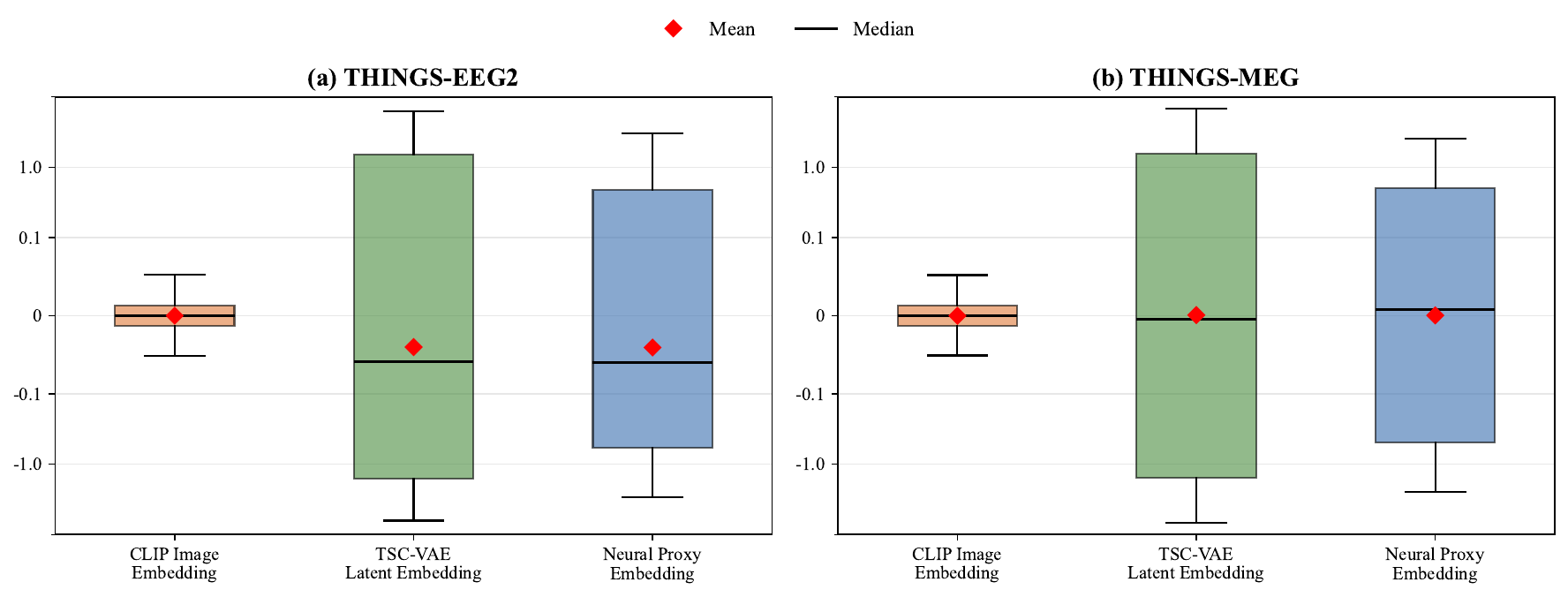}
\caption{Distribution comparison of CLIP image embeddings, TSC-VAE latent embeddings, and neural proxy embeddings on (a) THINGS-EEG2 and (b) THINGS-MEG under a symmetric log scale. The whiskers extend from the 1st to the 99th percentile, and the box spans the interquartile range (Q25--Q75). The red diamond denotes the mean and the black horizontal line denotes the median. Detailed statistics are reported in Table~\ref{tab_embedding_stats}.}
\label{fig_embedding_analysis}
\end{figure*}
\begin{table*}[htbp!]
\caption{Embedding distribution statistics averaged across all subjects.}
\label{tab_embedding_stats}
\centering
\begin{minipage}[t]{0.48\textwidth}
\centering
\textbf{(a) THINGS-EEG2} \\[0.5em]
\resizebox{\textwidth}{!}{
\begin{tabular}{c|ccc}
\toprule
& \makecell{CLIP Image\\Embedding} & \makecell{TSC-VAE\\Latent Embedding} & \makecell{Neural Proxy\\Embedding} \\
\midrule
Mean & $-$0.0001 & $-$0.0402 & $-$0.0408 \\
Std & 0.0247 & 2.5414 & 1.0465 \\
Median & $-$0.0002 & $-$0.0589 & $-$0.0603 \\
Q25 & $-$0.0134 & $-$1.6023 & $-$0.5812 \\
Q75 & 0.0131 & 1.5125 & 0.4744 \\
Q01 & $-$0.0511 & $-$6.3516 & $-$2.9357 \\
Q99 & 0.0522 & 6.2973 & 3.0320 \\
\bottomrule
\end{tabular}
}
\label{tab_eeg_embedding_stats}
\end{minipage}
\hfill
\begin{minipage}[t]{0.48\textwidth}
\centering
\textbf{(b) THINGS-MEG} \\[0.5em]
\resizebox{\textwidth}{!}{
\begin{tabular}{c|ccc}
\toprule
& \makecell{CLIP Image\\Embedding} & \makecell{TSC-VAE\\Latent Embedding} & \makecell{Neural Proxy\\Embedding} \\
\midrule
Mean & $-$0.0001 & 0.0009 & 0.0004 \\
Std & 0.0246 & 2.6688 & 0.9413 \\
Median & $-$0.0002 & $-$0.0045 & 0.0078 \\
Q25 & $-$0.0134 & $-$1.5534 & $-$0.4959 \\
Q75 & 0.0130 & 1.5485 & 0.5057 \\
Q01 & $-$0.0509 & $-$6.8126 & $-$2.5122 \\
Q99 & 0.0520 & 6.8643 & 2.5348 \\
\bottomrule
\end{tabular}
}
\label{tab_meg_embedding_stats}
\end{minipage}
\end{table*}
A natural question arises: how large is the modality gap between visual and neural embeddings, and does the Q-Former effectively bridge it?
To answer this, we analyze the distributional characteristics of three embedding types: CLIP image embeddings $\mathbf{e}_{\mathbf{V}}$, TSC-VAE latent embeddings $\mathbf{z}$, and neural proxy embeddings $\hat{\mathbf{z}}$, with statistics computed on the combined training and test sets and averaged across all subjects (Fig.~\ref{fig_embedding_analysis} and Table~\ref{tab_embedding_stats}).
A striking scale discrepancy exists between the CLIP image embeddings and the TSC-VAE latent embeddings.
On THINGS-EEG2, the CLIP image embeddings have a standard deviation of only 0.0247 with a 1st--99th percentile range of $[-0.0511,\, 0.0522]$, whereas the TSC-VAE latent embeddings exhibit a standard deviation of 2.5414 with a range of $[-6.3516,\, 6.2973]$---a roughly two orders of magnitude difference.
A similar pattern is observed on THINGS-MEG, where the CLIP standard deviation is 0.0246 compared to 2.6688 for the TSC-VAE latent embeddings.
This confirms that the visual semantic space occupied by CLIP image embeddings and the neural representation space of TSC-VAE latent embeddings lie on vastly different scales, making direct alignment infeasible.
After training the Q-Former with the alignment loss $\mathcal{L}_{\text{Align}}$ (Eq.~\ref{eq:align_loss}), the resulting neural proxy embeddings $\hat{\mathbf{z}}$ exhibit a substantially expanded scale that is much closer to that of the TSC-VAE latent embeddings.
On THINGS-EEG2, the neural proxy embeddings have a standard deviation of 1.0465 with a range of $[-2.9357,\, 3.0320]$; on THINGS-MEG, the standard deviation is 0.9413 with a range of $[-2.5122,\, 2.5348]$.
Compared to the original CLIP image embeddings, the scale of the neural proxy embeddings is approximately 40 times larger, bringing them significantly closer to the TSC-VAE latent space.
These results demonstrate that the Q-Former projection effectively bridges the modality gap by transforming compact visual semantic features into the broader neural representation scale, validating the design choice presented in Section~\ref{sec:stage2}.

\subsection{Loss Function Ablation Study}
\label{sec:ablation_loss}

\begin{table*}[htbp!]
\caption{Ablation study on loss functions for cross-modal alignment on THINGS-EEG2 and THINGS-MEG datasets.}
\label{tab_loss_ablation}
\centering
\begin{minipage}[t]{0.48\textwidth}
\centering
\textbf{(a) THINGS-EEG2} \\[0.5em]
\resizebox{\textwidth}{!}{
\begin{tabular}{c|ccc}
\toprule
\multirow{2}{*}{Loss Function} & \multicolumn{3}{c}{Evaluation Metrics} \\
\cmidrule{2-4}
\multirow{2}{*}{~} & MSE $\downarrow$ & Pearson $\uparrow$ & Cosine $\uparrow$ \\
\midrule
$\mathcal{L}_{\text{MSE}}$ & $\boldsymbol{0.828}$ & 0.621 & 0.620 \\
$\mathcal{L}_{\text{SWD}}$ & 1.012 & 0.611 & 0.612 \\
$\mathcal{L}_{\text{MSE}} + \mathcal{L}_{\text{SWD}}$ & 0.954 & $\boldsymbol{0.635}$ & $\boldsymbol{0.635}$ \\
\bottomrule
\end{tabular}
}
\label{tab_eeg_loss_ablation}
\end{minipage}
\hfill
\begin{minipage}[t]{0.48\textwidth}
\centering
\textbf{(b) THINGS-MEG} \\[0.5em]
\resizebox{\textwidth}{!}{
\begin{tabular}{c|ccc}
\toprule
\multirow{2}{*}{Loss Function} & \multicolumn{3}{c}{Evaluation Metrics} \\
\cmidrule{2-4}
\multirow{2}{*}{~} & MSE $\downarrow$ & Pearson $\uparrow$ & Cosine $\uparrow$ \\
\midrule
$\mathcal{L}_{\text{MSE}}$ & $\boldsymbol{1.315}$ & 0.537 & 0.533 \\
$\mathcal{L}_{\text{SWD}}$ & 1.504 & 0.530 & 0.529 \\
$\mathcal{L}_{\text{MSE}} + \mathcal{L}_{\text{SWD}}$ & 1.451 & $\boldsymbol{0.543}$ & $\boldsymbol{0.539}$ \\
\bottomrule
\end{tabular}
}
\label{tab_meg_loss_ablation}
\end{minipage}
\end{table*}

To evaluate the contribution of each loss component, we conduct an ablation study comparing $\mathcal{L}_{\text{MSE}}$ alone, $\mathcal{L}_{\text{SWD}}$ alone, and the combined $\mathcal{L}_{\text{MSE}} + \mathcal{L}_{\text{SWD}}$, as shown in Table~\ref{tab_loss_ablation}.
Using $\mathcal{L}_{\text{MSE}}$ alone achieves the lowest MSE on both datasets, while $\mathcal{L}_{\text{SWD}}$ alone yields competitive Pearson correlation and cosine similarity but with higher MSE.
Combining $\mathcal{L}_{\text{MSE}} + \mathcal{L}_{\text{SWD}}$ achieves the best Pearson correlation and cosine similarity on both datasets (0.635/0.635 on THINGS-EEG2 and 0.543/0.539 on THINGS-MEG), confirming that $\mathcal{L}_{\text{SWD}}$ provides complementary distributional alignment that $\mathcal{L}_{\text{MSE}}$ alone cannot capture.

\subsection{Brain Region Ablation Study}
\label{sec:brain_region_ablation_study}


%
\begin{figure*}[htbp!]
\centering
\begin{minipage}[t]{0.48\textwidth}
\centering
\includegraphics[width=\linewidth]{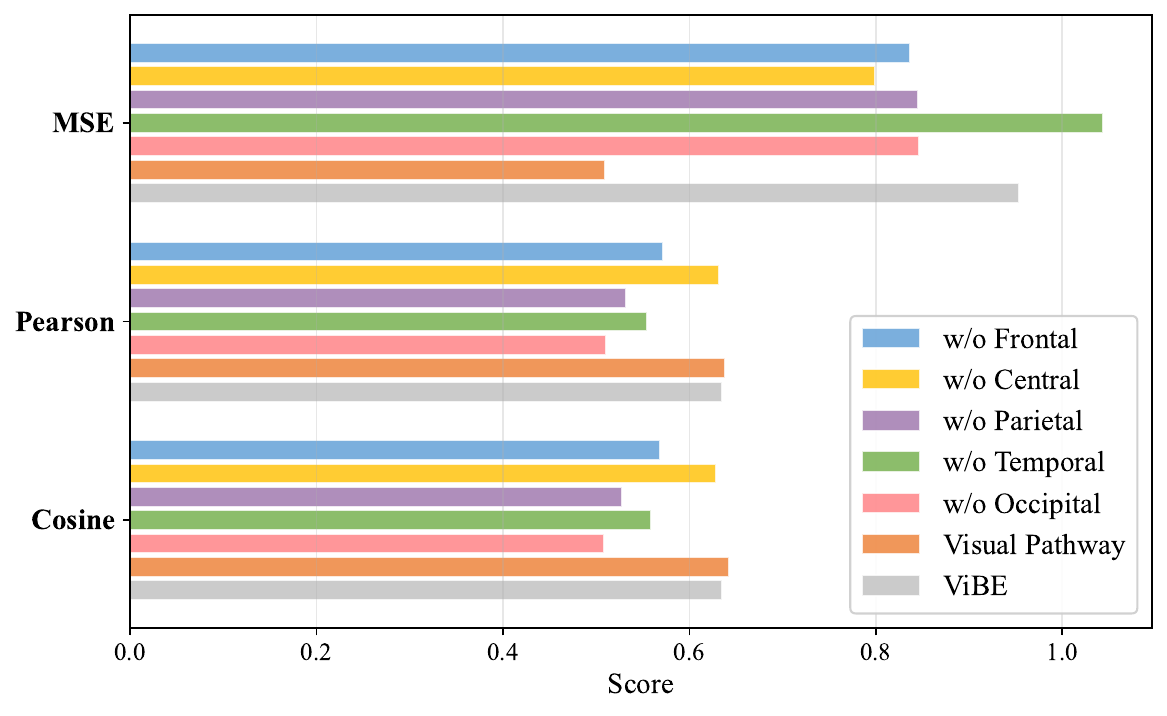}
\end{minipage}
\hfill
\begin{minipage}[t]{0.48\textwidth}
\centering
\includegraphics[width=\linewidth]{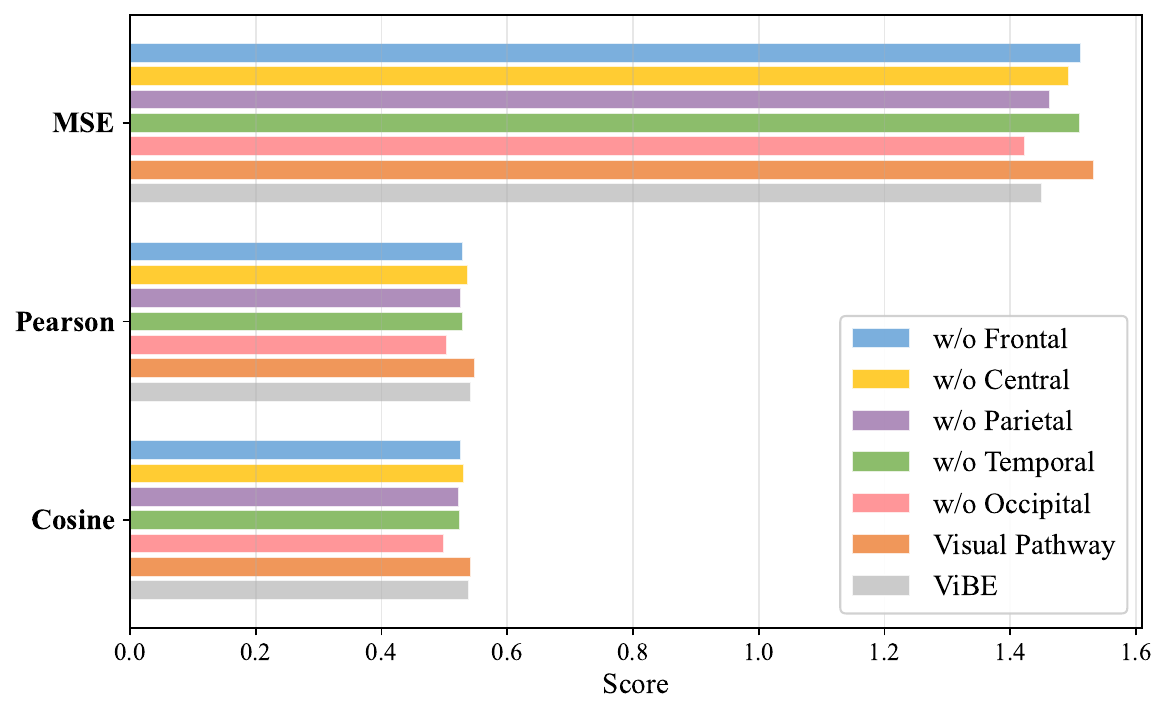}
\end{minipage}
\caption{
Brain region ablation study results. Left: THINGS-EEG2. Right: THINGS-MEG.
}
\label{fig_brain_region_ablation}
\end{figure*}
To investigate the contribution of different brain regions to brain signal generation, we conduct ablation studies by systematically removing channels from each brain region (the detailed brain region configurations are provided in Appendix~\ref{appendix:brain_region_config}).
Additionally, we design a Visual Pathway condition that retains only the temporal and occipital channels for training, following the bottom-up hierarchy of the visual system~\citep{song2023decoding}.
As shown in Figure~\ref{fig_brain_region_ablation}, across both datasets, the occipital region exhibits the strongest influence on performance.
Removing the occipital channels leads to the largest degradation in both Pearson correlation and cosine similarity among all ablation conditions.
This finding is consistent with the well-established role of the occipital cortex as the primary visual processing area, where V1, V2, and V4 are responsible for encoding visual stimuli~\citep{song2023decoding}.
Interestingly, the Visual Pathway condition achieves Pearson correlation and cosine similarity scores that are slightly higher than ViBE on both datasets.
This result provides strong empirical support for the ventral stream hypothesis, indicating that the temporal and occipital cortices carry the most informative visual representations.

\section{Conclusion}
\label{sec:conclusion}
In this paper, we propose ViBE, a novel brain encoding framework for generating M/EEG signals from visual stimuli.
Our framework comprises two stages: in Stage~I, the TSC-VAE captures the hierarchical spatio-temporal characteristics of M/EEG signals by employing temporal-spatial separated convolutions with reduced spatial kernel sizes, achieving high-fidelity neural response reconstruction.
In Stage~II, Q-Former maps CLIP image embeddings to the TSC-VAE latent space, producing neural proxy embeddings that are aligned with TSC-VAE latent embeddings through a combined MSE and SWD loss for both point-wise and distributional alignment.
Extensive experiments on the THINGS-EEG2 and THINGS-MEG datasets demonstrate the effectiveness of our approach.

\bibliographystyle{unsrt}
\bibliography{references}

\newpage
\appendix

\section{Technical appendices and supplementary material}
\subsection{Experiment Details}
\label{appendix:experiment_details}

We implement ViBE using PyTorch on four NVIDIA V100S GPUs.
\textbf{Datasets and Preprocessing.}
We evaluate on THINGS-EEG2~\cite{gifford2022large} and THINGS-MEG~\cite{hebart2023things}.
THINGS-EEG2 contains EEG recordings from 10 subjects using 63 electrode channels at 1000\,Hz. The training set comprises 1,654 concepts $\times$ 10 images $\times$ 4 repetitions per subject, and the test set includes 200 concepts $\times$ 1 image $\times$ 80 repetitions. The data is epoched from $-200$\,ms to 1000\,ms relative to stimulus onset, baseline corrected, downsampled to 250\,Hz, and whitened using Multivariate Noise Normalization (MVNN)~\cite{guggenmos2018multivariate}. The final EEG data has dimensions of $63 \times 250$.
THINGS-MEG contains MEG recordings from 4 participants using 271 magnetometer channels.
The training set comprises 1,854 concepts $\times$ 12 images $\times$ 1 repetition per participant, and the test set includes 200 concepts $\times$ 1 image $\times$ 12 repetitions.
The data is bandpass filtered at [0.1, 100]\,Hz, epoched from $-100$\,ms to 1300\,ms, baseline corrected using z-score normalization, cropped to 0--1000\,ms, and downsampled to 200\,Hz. The final MEG data has dimensions of $271 \times 200$.
\textbf{Stage~I: Neural Response Reconstruction via TSC-VAE.}
The TSC-VAE encoder employs temporal-spatial separated convolutions with TSConvResBlocks.
The encoder consists of an initial temporal convolution layer (kernel size $1 \times 25$, mapping 1 input channel to 32 feature channels), followed by alternating TSConvResBlocks (with temporal kernel sizes of 15, 11, 7 and spatial kernel size of 3) and downsampling blocks (stride 2), mapping the input signal to a latent space of dimension $d \times H \times W$ with $d = 4$ latent channels.
For EEG, the latent shape is $4 \times 16 \times 63$; for MEG, $4 \times 68 \times 50$.
The decoder mirrors the encoder architecture with upsampling blocks.
We train the TSC-VAE using the AdamW optimizer with a learning rate of $1 \times 10^{-4}$, weight decay of $1 \times 10^{-5}$, and cosine annealing learning rate schedule.
The KL divergence weight $\beta = 1 \times 10^{-4}$ is linearly warmed up over the first 10 epochs.
The batch size is 64 for EEG and 16 for MEG, and training runs for 100 epochs.
\textbf{Stage~II: Visual-to-Neural Mapping via Q-Former Projection.}
The Q-Former module uses 64 learnable query tokens with a hidden dimension of 768, 6 transformer layers, 8 attention heads, and an MLP ratio of 4.0.
Cross-attention layers are inserted every 2 layers.
The CLIP visual encoder (ViT-L/14) produces 257 visual tokens (1 CLS token + 256 patch tokens) with dimension 768, which serve as the key and value in cross-attention.
Both the CLIP encoder and the TSC-VAE are frozen during Stage~II training.
We use the AdamW optimizer with a learning rate of $1 \times 10^{-4}$, weight decay of $1 \times 10^{-5}$, gradient clipping at 1.0, and a cosine annealing schedule with 5 epochs of linear warmup.
The SWD loss uses 50 random projections along the channel dimension.
Training runs for 100 epochs with the same batch sizes as Stage~I.
%

\subsection{Evaluation Protocols}
\label{appendix:eval_protocols}

THINGS-EEG2 contains data from 10 subjects and THINGS-MEG from 4 subjects, each split into training and test sets.
For the ``ViBE (Ours)'' setting, we train a separate model for each subject using its own training data and evaluate on its test data; the reported results are averaged across all subjects.
For the ``ViBE (Ours, cross-subject)'' setting, each per-subject model is evaluated on the test data of all other subjects.
We additionally adopt a leave-one-subject-out protocol to mitigate cross-subject performance degradation.
Taking THINGS-EEG2 as an example: for each fold, we hold out one subject and combine the training data of the remaining 9 subjects into a unified training set, with their test data forming a unified test set.
A single model is trained on this combined training set and then evaluated in two ways: (1) on the combined test set of the 9 training subjects, yielding the ``Within-subject'' result, and (2) on the test data of the held-out subject, yielding the ``Cross-subject'' result.
Averaging across all 10 folds produces the final leave-one-subject-out results.
The same protocol is applied to THINGS-MEG with 4 subjects.

\subsection{Analysis of MSE versus Pearson/Cosine Discrepancy}
\label{appendix:mse_analysis}

As shown in Table~\ref{tab_results}, ViBE achieves substantially higher Pearson correlation and cosine similarity than all baselines on both datasets, yet exhibits higher MSE.
This is a consequence of the residual amplitude scale offset between the neural proxy embeddings and the TSC-VAE latent embeddings: as reported in Table~\ref{tab_embedding_stats}, the neural proxy embeddings have standard deviations of approximately 1.0 (1.0465 on THINGS-EEG2, 0.9413 on THINGS-MEG), while the TSC-VAE latent embeddings have standard deviations of approximately 2.5 (2.5414 on THINGS-EEG2, 2.6688 on THINGS-MEG)---a residual scale gap of roughly 2.5$\times$.
Since MSE computes the point-wise squared difference and is sensitive to absolute amplitude, this systematic attenuation in the generated signals inflates the MSE, even though the waveform patterns faithfully match the ground truth neural responses (yielding high Pearson correlation and cosine similarity).
Furthermore, the loss ablation results in Table~\ref{tab_loss_ablation} provide additional evidence. Using $\mathcal{L}_{\text{SWD}}$ alone yields the highest MSE on both datasets (1.012 on THINGS-EEG2 and 1.504 on THINGS-MEG), because $\mathcal{L}_{\text{SWD}}$ performs probability distribution alignment rather than point-wise optimization, and therefore does not directly minimize point-wise differences.
When combining $\mathcal{L}_{\text{MSE}} + \mathcal{L}_{\text{SWD}}$, the MSE increases compared to using $\mathcal{L}_{\text{MSE}}$ alone (0.828 $\rightarrow$ 0.954 on THINGS-EEG2 and 1.315 $\rightarrow$ 1.451 on THINGS-MEG), as the distributional alignment objective of $\mathcal{L}_{\text{SWD}}$ introduces a trade-off with the point-wise minimization of $\mathcal{L}_{\text{MSE}}$.
However, this trade-off is beneficial: the combined loss achieves the best Pearson correlation and cosine similarity, confirming that $\mathcal{L}_{\text{SWD}}$ provides complementary distributional alignment that $\mathcal{L}_{\text{MSE}}$ alone cannot capture.

\subsection{Brain Region Configuration}
\label{appendix:brain_region_config}

This appendix provides the detailed brain region configurations used in the brain region ablation study (Section~\ref{sec:brain_region_ablation_study}).
\textbf{EEG Configuration.}
For THINGS-EEG2 with 63 electrode channels, we define 5 brain regions based on the standard 10--10 electrode naming system:
\begin{itemize}
    \item Frontal (22 channels): Fp1, Fp2, AF7, AF3, AFz, AF4, AF8, F7, F5, F3, F1, F2, F4, F6, F8, FC5, FC3, FC1, FCz, FC2, FC4, FC6
    \item Central (14 channels): C5, C3, C1, Cz, C2, C4, C6, CP5, CP3, CP1, CPz, CP2, CP4, CP6
    \item Temporal (10 channels): FT9, FT7, FT8, FT10, T7, T8, TP9, TP7, TP8, TP10
    \item Parietal (14 channels): P7, P5, P3, P1, Pz, P2, P4, P6, P8, PO7, PO3, POz, PO4, PO8
    \item Occipital (3 channels): O1, Oz, O2
\end{itemize}
\textbf{MEG Configuration.}
For THINGS-MEG with 271 magnetometer channels (after excluding one faulty channel), we define 5 brain regions based on the CTF channel naming convention, where each channel name follows the format M[L/R/Z][F/C/P/O/T][number] (L = left, R = right, Z = midline; F = frontal, C = central, P = parietal, O = occipital, T = temporal):
\begin{itemize}
    \item Frontal (67 channels): 32 left (MLF), 32 right (MRF), 3 midline (MZF)
    \item Central (52 channels): 24 left (MLC), 24 right (MRC), 4 midline (MZC)
    \item Parietal (45 channels): 22 left (MLP), 22 right (MRP), 1 midline (MZP)
    \item Occipital (39 channels): 19 left (MLO), 17 right (MRO), 3 midline (MZO)
    \item Temporal (68 channels): 34 left (MLT), 34 right (MRT)
\end{itemize}
For both EEG and MEG, the Visual Pathway condition retains only the temporal and occipital channels (13 channels for EEG, 107 channels for MEG), following the bottom-up hierarchy of the visual system along the ventral stream~\cite{song2023decoding}.

\subsection{Training and Inference Algorithms}
\label{appendix:algorithms}

This appendix presents the pseudocode for the two-stage training procedure and the inference pipeline of ViBE.
The entire process proceeds sequentially: we first train the TSC-VAE (Algorithm~\ref{alg:stage1}) to learn a compact latent representation of M/EEG signals, then train the Q-Former (Algorithm~\ref{alg:stage2}) to map CLIP image embeddings into this learned latent space while keeping both the CLIP encoder and the TSC-VAE encoder frozen, and finally perform inference (Algorithm~\ref{alg:inference}) by composing all trained components into a single forward pass.

\begin{algorithm}[H]
\caption{Stage~I: TSC-VAE Training}
\label{alg:stage1}
\begin{algorithmic}[1]
\Require Training dataset $\mathcal{D} = \{(\mathbf{V}^{(i)}, \mathbf{x}^{(i)})\}_{i=1}^{N}$, KL weight $\beta$, warmup epochs $E_w$, total epochs $E$
\Ensure Trained TSC-VAE with encoder $\mathcal{E}_\theta$ and decoder $\mathcal{D}_\phi$
\State Initialize $\mathcal{E}_\theta$ and $\mathcal{D}_\phi$
\For{epoch $= 1$ to $E$}
    \State $\beta_{\text{cur}} \leftarrow \beta \cdot \min(1,\; \text{epoch} / E_w)$ \Comment{KL warmup}
    \For{each mini-batch $\{\mathbf{x}\}$}
        \State $(\boldsymbol{\mu}, \log \boldsymbol{\sigma}^2) \leftarrow \mathcal{E}_\theta(\mathbf{x})$
        \State $\boldsymbol{\epsilon} \sim \mathcal{N}(\mathbf{0}, \mathbf{I})$
        \State $\mathbf{z} \leftarrow \boldsymbol{\mu} + \boldsymbol{\epsilon} \odot \boldsymbol{\sigma}$ \Comment{Reparameterization}
        \State $\hat{\mathbf{x}} \leftarrow \mathcal{D}_\phi(\mathbf{z})$
        \State $\mathcal{L}_{\text{VAE}} \leftarrow \text{MSE}(\hat{\mathbf{x}}, \mathbf{x}) + \beta_{\text{cur}} \cdot D_{\text{KL}}(q_\theta(\mathbf{z}|\mathbf{x}) \| p(\mathbf{z}))$
        \State Update $\theta, \phi$ via gradient descent on $\mathcal{L}_{\text{VAE}}$
    \EndFor
\EndFor
\State \Return $\mathcal{E}_\theta$, $\mathcal{D}_\phi$
\end{algorithmic}
\end{algorithm}

\begin{algorithm}[H]
\caption{Stage~II: Q-Former Training}
\label{alg:stage2}
\begin{algorithmic}[1]
\Require Training dataset $\mathcal{D}$, frozen CLIP encoder $f_{\text{CLIP}}$, frozen TSC-VAE encoder $\mathcal{E}_\theta$, total epochs $E$
\Ensure Trained Q-Former $g_\psi$ with projection head $P$
\State Initialize Q-Former $g_\psi$ and projection head $P$
\For{epoch $= 1$ to $E$}
    \For{each mini-batch $\{(\mathbf{V}, \mathbf{x})\}$}
        \State $\mathbf{e}_{\mathbf{V}} \leftarrow f_{\text{CLIP}}(\mathbf{V})$ \Comment{CLIP image embeddings (frozen)}
        \State $\mathbf{z} \leftarrow \mathcal{E}_\theta(\mathbf{x})$ \Comment{Target latent (frozen)}
        \State $\hat{\mathbf{z}} \leftarrow P(g_\psi(\mathbf{e}_{\mathbf{V}}))$ \Comment{Neural proxy embeddings}
        \State $\mathcal{L}_{\text{MSE}} \leftarrow \text{MSE}(\hat{\mathbf{z}}, \mathbf{z})$
        \State $\mathcal{L}_{\text{SWD}} \leftarrow \frac{1}{M}\sum_{m=1}^{M} W_1(\boldsymbol{\theta}_m^\top \hat{\mathbf{z}}_{\text{flat}},\; \boldsymbol{\theta}_m^\top \mathbf{z}_{\text{flat}})$
        \State $\mathcal{L}_{\text{Align}} \leftarrow \mathcal{L}_{\text{MSE}} + \lambda \mathcal{L}_{\text{SWD}}$
        \State Update $\psi, P$ via gradient descent on $\mathcal{L}_{\text{Align}}$
    \EndFor
\EndFor
\State \Return $g_\psi$, $P$
\end{algorithmic}
\end{algorithm}

\begin{algorithm}[H]
\caption{Inference}
\label{alg:inference}
\begin{algorithmic}[1]
\Require Test image $\mathbf{V}$, frozen CLIP encoder $f_{\text{CLIP}}$, trained Q-Former $g_\psi$ with projection head $P$, frozen TSC-VAE decoder $\mathcal{D}_\phi$
\Ensure Predicted M/EEG signal $\hat{\mathbf{x}}$
\State $\mathbf{e}_{\mathbf{V}} \leftarrow f_{\text{CLIP}}(\mathbf{V})$ \Comment{Extract CLIP image embeddings}
\State $\hat{\mathbf{z}} \leftarrow P(g_\psi(\mathbf{e}_{\mathbf{V}}))$ \Comment{Produce neural proxy embeddings}
\State $\hat{\mathbf{x}} \leftarrow \mathcal{D}_\phi(\hat{\mathbf{z}})$ \Comment{Decode to M/EEG signal}
\State \Return $\hat{\mathbf{x}}$
\end{algorithmic}
\end{algorithm}

\subsection{Detailed Per-Subject Results}
\label{appendix:detailed_results}

Tables~\ref{tab_eeg_detailed} and \ref{tab_meg_detailed} present the detailed per-subject evaluation results corresponding to Table~\ref{tab_results} in the main text.
Stage~I reports the reconstruction performance of TSC-VAE, reflecting how faithfully the autoencoder can reconstruct the original M/EEG signals.
Stage~II reports the end-to-end encoding performance of the full pipeline (from visual stimuli to predicted M/EEG).
Cross-subject denotes the cross-subject generalization setting where the model trained on one subject is directly evaluated on other subjects.
\begin{table*}[htbp!]
\caption{Detailed per-subject results on THINGS-EEG2 (10 subjects).}
\label{tab_eeg_detailed}
\centering
\resizebox{\textwidth}{!}{
\begin{tabular}{cc|cccccccccc|c}
\toprule
\multicolumn{2}{c|}{} & Sub 1 & Sub 2 & Sub 3 & Sub 4 & Sub 5 & Sub 6 & Sub 7 & Sub 8 & Sub 9 & Sub 10 & AVG \\
\midrule
\multicolumn{13}{l}{\textit{Subject-specific}} \\
\cmidrule{1-13}
\multirow{3}{*}{MSE $\downarrow$} & Stage I & 0.0039 & 0.0046 & 0.0064 & 0.0029 & 0.0045 & 0.0045 & 0.0028 & 0.0041 & 0.0044 & 0.0047 & 0.0043 \\
& Stage II & 0.9748 & 0.8717 & 1.3957 & 0.8575 & 0.6822 & 0.6044 & 1.1352 & 0.9406 & 1.1843 & 0.8903 & 0.9537 \\
& Cross-subject & 1.7375 & 1.7764 & 2.7790 & 1.6333 & 1.5840 & 1.4750 & 2.1529 & 1.7109 & 1.7588 & 2.0376 & 1.8645 \\
\cmidrule{1-13}
\multirow{3}{*}{Pearson $\uparrow$} & Stage I & 0.9545 & 0.9577 & 0.8953 & 0.9702 & 0.9388 & 0.9269 & 0.9675 & 0.9491 & 0.9082 & 0.9439 & 0.9412 \\
& Stage II & 0.6165 & 0.6482 & 0.6785 & 0.6522 & 0.5924 & 0.6197 & 0.6410 & 0.7214 & 0.5199 & 0.6621 & 0.6352 \\
& Cross-subject & 0.2898 & 0.3341 & 0.2389 & 0.4024 & 0.3068 & 0.2737 & 0.2597 & 0.3187 & 0.2279 & 0.2936 & 0.2946 \\
\cmidrule{1-13}
\multirow{3}{*}{Cosine $\uparrow$} & Stage I & 0.9532 & 0.9578 & 0.8967 & 0.9704 & 0.9399 & 0.9267 & 0.9673 & 0.9494 & 0.9093 & 0.9445 & 0.9415 \\
& Stage II & 0.6154 & 0.6465 & 0.6808 & 0.6531 & 0.5924 & 0.6136 & 0.6407 & 0.7222 & 0.5243 & 0.6599 & 0.6349 \\
& Cross-subject & 0.2899 & 0.3336 & 0.2445 & 0.4025 & 0.3106 & 0.2692 & 0.2602 & 0.3241 & 0.2366 & 0.2939 & 0.2965 \\
\midrule
\multicolumn{13}{l}{\textit{Leave-one-subject-out}} \\
\cmidrule{1-13}
\multirow{3}{*}{MSE $\downarrow$} & Stage I & 0.0034 & 0.0033 & 0.0033 & 0.0034 & 0.0033 & 0.0032 & 0.0035 & 0.0034 & 0.0033 & 0.0033 & 0.0033 \\
& Within-subject & 0.3279 & 0.3390 & 0.3892 & 0.3672 & 0.3774 & 0.3838 & 0.3839 & 0.3437 & 0.3804 & 0.3271 & 0.3620 \\
& Cross-subject & 0.3374 & 0.3613 & 0.4424 & 0.3642 & 0.4208 & 0.4512 & 0.4149 & 0.3480 & 0.4303 & 0.3864 & 0.3957 \\
\cmidrule{1-13}
\multirow{3}{*}{Pearson $\uparrow$} & Stage I & 0.9546 & 0.9548 & 0.9560 & 0.9525 & 0.9558 & 0.9577 & 0.9518 & 0.9533 & 0.9592 & 0.9564 & 0.9552 \\
& Within-subject & 0.4505 & 0.4736 & 0.4908 & 0.4716 & 0.4785 & 0.4905 & 0.4826 & 0.4703 & 0.5004 & 0.4869 & 0.4796 \\
& Cross-subject & 0.4311 & 0.5058 & 0.3797 & 0.5176 & 0.4412 & 0.3807 & 0.4508 & 0.5563 & 0.3288 & 0.4544 & 0.4446 \\
\cmidrule{1-13}
\multirow{3}{*}{Cosine $\uparrow$} & Stage I & 0.9551 & 0.9551 & 0.9565 & 0.9531 & 0.9563 & 0.9579 & 0.9525 & 0.9532 & 0.9595 & 0.9562 & 0.9555 \\
& Within-subject & 0.4513 & 0.4746 & 0.4891 & 0.4713 & 0.4787 & 0.4906 & 0.4809 & 0.4702 & 0.5007 & 0.4868 & 0.4794 \\
& Cross-subject & 0.4318 & 0.5056 & 0.3786 & 0.5158 & 0.4416 & 0.3805 & 0.4504 & 0.5566 & 0.3306 & 0.4546 & 0.4446 \\
\bottomrule
\end{tabular}
}
\end{table*}
\begin{table*}[htbp!]
\caption{Detailed per-subject results on THINGS-MEG (4 subjects).}
\label{tab_meg_detailed}
\centering
\resizebox{0.7\textwidth}{!}{
\begin{tabular}{cc|cccc|c}
\toprule
\multicolumn{2}{c|}{} & Sub 1 & Sub 2 & Sub 3 & Sub 4 & AVG \\
\midrule
\multicolumn{7}{l}{\textit{Subject-specific}} \\
\cmidrule{1-7}
\multirow{3}{*}{MSE $\downarrow$} & Stage I & 0.0122 & 0.0245 & 0.0375 & 0.0138 & 0.0220 \\
& Stage II & 1.0793 & 1.5353 & 1.8672 & 1.3221 & 1.4510 \\
& Cross-subject & 1.6910 & 2.1828 & 2.8599 & 1.8802 & 2.1535 \\
\cmidrule{1-7}
\multirow{3}{*}{Pearson $\uparrow$} & Stage I & 0.9850 & 0.9822 & 0.9740 & 0.9819 & 0.9808 \\
& Stage II & 0.5734 & 0.6291 & 0.4973 & 0.4730 & 0.5432 \\
& Cross-subject & 0.2985 & 0.3077 & 0.2028 & 0.2356 & 0.2612 \\
\cmidrule{1-7}
\multirow{3}{*}{Cosine $\uparrow$} & Stage I & 0.9851 & 0.9821 & 0.9740 & 0.9816 & 0.9807 \\
& Stage II & 0.5724 & 0.6262 & 0.4942 & 0.4640 & 0.5392 \\
& Cross-subject & 0.2985 & 0.3064 & 0.2029 & 0.2338 & 0.2604 \\
\midrule
\multicolumn{7}{l}{\textit{Leave-one-subject-out}} \\
\cmidrule{1-7}
\multirow{3}{*}{MSE $\downarrow$} & Stage I & 0.0242 & 0.0198 & 0.0156 & 0.0216 & 0.0203 \\
& Within-subject & 1.5459 & 1.3010 & 0.7047 & 1.3911 & 1.2357 \\
& Cross-subject & 1.5808 & 1.5077 & 1.0037 & 1.5690 & 1.4153 \\
\cmidrule{1-7}
\multirow{3}{*}{Pearson $\uparrow$} & Stage I & 0.9809 & 0.9814 & 0.9842 & 0.9818 & 0.9821 \\
& Within-subject & 0.4259 & 0.4177 & 0.4704 & 0.4832 & 0.4493 \\
& Cross-subject & 0.4033 & 0.3761 & 0.2637 & 0.2896 & 0.3332 \\
\cmidrule{1-7}
\multirow{3}{*}{Cosine $\uparrow$} & Stage I & 0.9793 & 0.9805 & 0.9844 & 0.9819 & 0.9815 \\
& Within-subject & 0.4226 & 0.4144 & 0.4679 & 0.4817 & 0.4467 \\
& Cross-subject & 0.4032 & 0.3726 & 0.2631 & 0.2868 & 0.3314 \\
\bottomrule
\end{tabular}
}
\end{table*}

\subsection{Limitations}
\label{appendix:limitations}
The publicly available M/EEG brain encoding datasets used in this work are limited in scale: THINGS-EEG2 contains only 10 subjects and THINGS-MEG contains only 4 subjects. This constrains the validation of our model in larger-scale settings. Future work will explore larger and more diverse datasets to further validate and improve the generalization capability of ViBE.
%


\end{document}